\documentclass{article}

    \usepackage[final]{tackling_climate_workshop_style}

\usepackage[utf8]{inputenc} % allow utf-8 input
\usepackage[T1]{fontenc}    % use 8-bit T1 fonts
\usepackage{hyperref}       % hyperlinks
\usepackage{url}            % simple URL typesetting
\usepackage{booktabs}       % professional-quality tables
\usepackage{amsfonts}       % blackboard math symbols
\usepackage{nicefrac}       % compact symbols for 1/2, etc.
\usepackage{microtype}      % microtypography

\usepackage{graphicx}       % insert figures

\title{Detecting Methane Plumes using PRISMA: Deep Learning Model and Data Augmentation}

\author{%
  Alexis Groshenry\\
  Kayrros SAS, Paris, France\\
  ENS Paris-Saclay, Paris, France\\
  \texttt{a.groshenry@kayrros.com}
  \And
  Clement Giron \\
  Kayrros SAS, Paris, France \\
  \texttt{c.giron@kayrros.com}
  \And
  Thomas Lauvaux \\
  University of Reims Champagne Ardenne, GSMA, UMR 7331, France \\
    \texttt{thomas.lauvaux@univ-reims.fr}
  \And
  Alexandre d'Aspremont \\
  CNRS, DI, Ecole Normale Supérieure \\
  Kayrros SAS, Paris, France \\
  \texttt{aspremon@ens.fr}
  \And
  Thibaud Ehret \\
  ENS Paris-Saclay, Paris, France \\
  \texttt{thibaud.ehret@ens-paris-saclay.fr}
}

\begin{document}

\maketitle

\begin{abstract}
The new generation of hyperspectral imagers, such as PRISMA, has improved significantly our detection capability of methane (CH$_4$) plumes from space at high spatial resolution ($\sim$30m). We present here a complete framework to identify CH$_4$ plumes using images from the PRISMA satellite mission and a deep learning model able to detect plumes over large areas. To compensate for the relative scarcity of PRISMA images, we trained our model by transposing high resolution plumes from Sentinel-2 to PRISMA. Our methodology thus avoids computationally expensive synthetic plume generation from Large Eddy Simulations by generating a broad and realistic training database, and paves the way for large-scale detection of methane plumes using future hyperspectral sensors (EnMAP, EMIT, CarbonMapper).
\end{abstract}

\section{Introduction}
Since 2019, the new generation of hyperspectral satellites (PRISMA, EnMAP, ...) collects an unprecedented amount of atmospheric data enabling the retrieval of methane (CH$_4$) concentrations over the globe at high spatial resolutions (around 30m per pixel). These satellite missions offer high resolution and broad spectral coverage, in particular in the shortwave infrared (1 to 2.5$\mu$m) where methane absorption is significant. 
These specifications enable lower detection thresholds compared to multispectral satellites for the identification and attribution of methane emissions from human activities (large releases from point sources), thus offering a path to reduce CH$_4$ emissions as pledged by nations engaged in climate mitigation agreements~\cite{IntergovernmentalPanelonClimateChange2021}.
% This makes possible the discovery of unknown leaks or faulty infrastructures and leads to better estimations of the corresponding emissions, thus offering a path to reduce CH$_4$ emissions as pledged by nations engaged in climate mitigation agreements~\cite{IntergovernmentalPanelonClimateChange2021}.
% These specifications make these satellites perfectly suited for the detection and attribution of methane emissions from human activities (large releases from point sources), and offer a path to reduce CH$_4$ emissions as pledged by nations engaged in climate mitigation agreements~\cite{IntergovernmentalPanelonClimateChange2021}.

The satellite PRISMA\footnote{Project carried out using PRISMA Products, © of the Italian Space Agency (ASI), delivered under an ASI Licence to Use} was launched in 2019 with forthcoming satellite missions like EnMAP offering similar coverage and performances. We propose here a complete processing pipeline for methane plumes identification on images from PRISMA, allowing efficient large-scale monitoring of these emissions. We present a novel automatic detection procedure that benefits from existing datasets of CH$_4$ images, collected by the previous generation of satellites (e.g. Sentinel). This approach can combine multiple spaceborne CH$_4$ imagers and enables the rapid development of a functional processing chain applicable to any future satellite instruments.

\section{Methods}

\subsection{Spectral Recalibration}

PRISMA is a satellite orbiting the Earth in a near polar, sun-synchronous trajectory \cite{prisma}. It carries a hyperspectral pushbroom sensor, made up of a line of 1,000 hyperspectral individual sensors scanning the ground.
% These individual sensors measure radiances in the shortwave infrared spectral domain $[900, 2500]$ nm, producing datacubes containing 171 spectral bands, 1,000 pixels in the across-track direction (one for each individual cell), and 1,000 pixels in the along-track direction.
% The spectral sampling of the satellite is 8 nm on average, with a mean FWHM of 16 nm.
A spectral calibration of the 1,000 hyperspectral sensors is provided with the datacubes. However, the nominal parameters used for the spectral calibration are not perfectly accurate since they were determined on ground, while heat perturbations or vibrations may alter the correctness of these parameters. Hence it is necessary to apply spectral recalibration to each PRISMA scene before further processing.

This preprocessing step is done by generating a theoretical radiance spectrum using a radiative transfer simulation modeling the emission of radiation by the Sun, its interactions with the Earth's atmosphere and reflection on the ground before being sent back towards the satellite's sensors. This is the so-called top-of-atmosphere radiance, which is eventually convolved with the sensors' parameters to get the simulated at-sensor radiance. The recalibration is then done by minimizing a distance criterion between the theoretical spectrum and the averaged observed spectrum for each sensor.

\subsection{Methane Retrieval}

We use the matched filter algorithm to derive the map of methane concentrations from raw radiance data from PRISMA. This algorithm allows for supervised target detection in signal processing and has been frequently used to produce gas concentrations retrieval \cite{prisma_guanter, thorpe_mf, thompson_2015}. In our study, we use the same formulation as described in \cite{prisma_guanter}.

For a target signal $t \in \mathbb{R}^d$ retrieved from an observed signal $x \in \mathbb{R}^d$ of mean $\mu$ and covariance $\Sigma$, the matched filter operator writes $\alpha(x) = \frac{(x-\mu)^T\Sigma^{-1}t}{t^T\Sigma^{-1}t}$. The observed signal $x$ is the raw radiance data and its statistics are computed for each sensor separately because of the differences of calibration. The target signal $t$ is defined as $\mu.k$, with $k$ being the Jacobian of the absorption spectrum for an additional ppm of methane, calculated using radiative transfer simulations based on the HITRAN database \cite{hitran} and tools from the LOWTRAN program \cite{lowtran}.

There are as many matched filters as sensors in the satellite's detector, combined to generate a map of column averaged mole fractions of CH$_4$ (XCH$_4$)\footnote{concentration of CH$_4$ in the column of atmosphere between the satellite's sensor and a pixel on the ground} on which the detection of methane plumes will be done.

\subsection{Automatic Plume Detection}

\subsubsection{Data}\label{data}
\begin{figure}[h!]
\begin{center}
\begin{minipage}{0.45\linewidth}
\centering \includegraphics[width = 7.5cm]{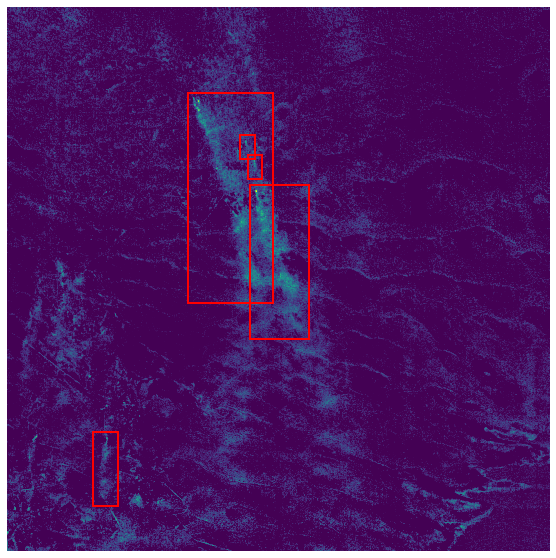}
\end{minipage}
\hfill
\begin{minipage}{0.45\linewidth}
\centering \includegraphics[width = 7.5cm]{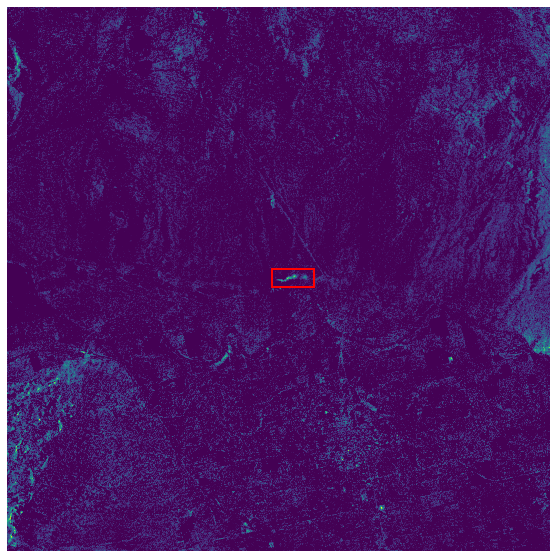}
\end{minipage}
\hfill
\begin{minipage}{0.45\linewidth}
\centering \includegraphics[width = 7.5cm]{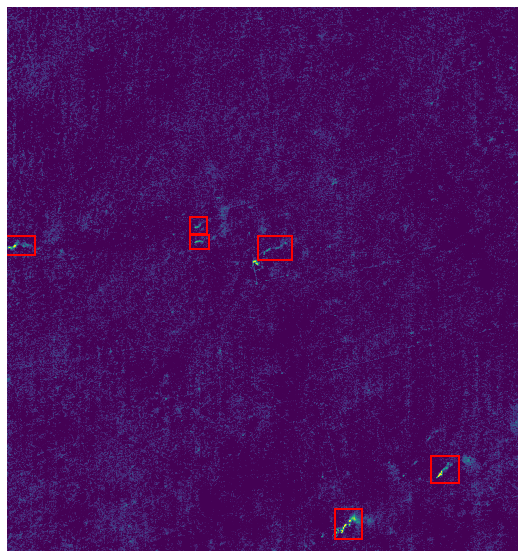}
\end{minipage}
\hfill
\begin{minipage}{0.45\linewidth}
\centering \includegraphics[width = 7.5cm]{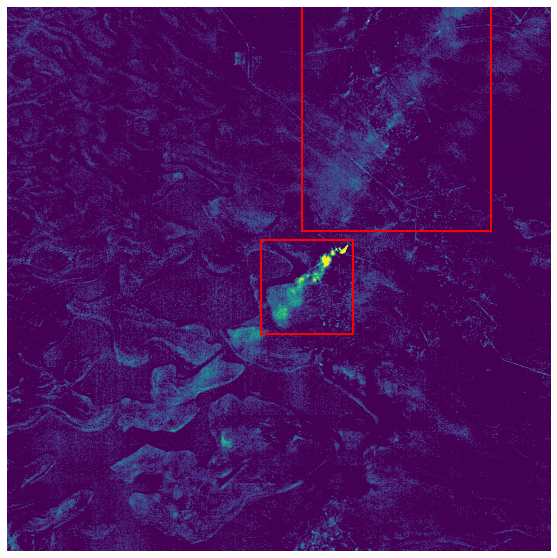}
\end{minipage}
\end{center}
\caption{\centering Some methane concentration maps illustrating the diversity of studied data. The bounding boxes correspond to manually identified methane plumes.}
\label{plume_examples}
\end{figure}

The main limitation in plume detection using PRISMA XCH$_4$ images is due to the small number of available images. PRISMA acquires data in tasking mode, hence producing few datacubes on targeted and localized areas, a limited proportion of which are relevant to our methane plume detection goal. Our study is based on 40 PRISMA images containing a total of 75 plumes of methane. This hardly describes the great diversity of plumes (i.e. size, intensity, morphology) and associated background (e.g. different levels of homogeneity, variable amount of noise, presence of clouds, roads, or buildings, types of terrain). To illustrate the diversity of observed scenes, we provide some examples in Fig~\ref{plume_examples}. In previous studies, several methods have been proposed to generate synthetic XCH$_4$ plumes, in addition to regular data augmentation techniques. Gaussian plume simulations \cite{bovensmann, krings} rely on a simple modelling of gas dispersion, but the final plume has a relatively naive shape, not representative of complex spatial structures observed from space. To simulate more complex structures, Large Eddy Simulations (LES) \cite{methanet, jongaramrungruang} rely on an accurate physical modelling of the atmospheric dispersion and turbulence to generate realistic plumes. But these models remain computationally expensive.

We propose here a novel synthetic plume generation technique that avoids the complex modelling of LES, while providing realistic plume shapes. The task of domain adaptation aims at changing the representation of a data, and seeks to transfer observed plume structures from one observing system to another. Inspired by this broad concept, we designed a plume transfer method from Sentinel-2 images to PRISMA, able to modify the inner distribution of Sentinel-2 XCH$_4$ plumes to match the characteristics of the PRISMA instrument, while preserving the shape and aspect of the original plume.

\begin{figure}[!h]
    \centering
    \makebox[0pt]{
    \centering
    \includegraphics[width=1.5\textwidth]{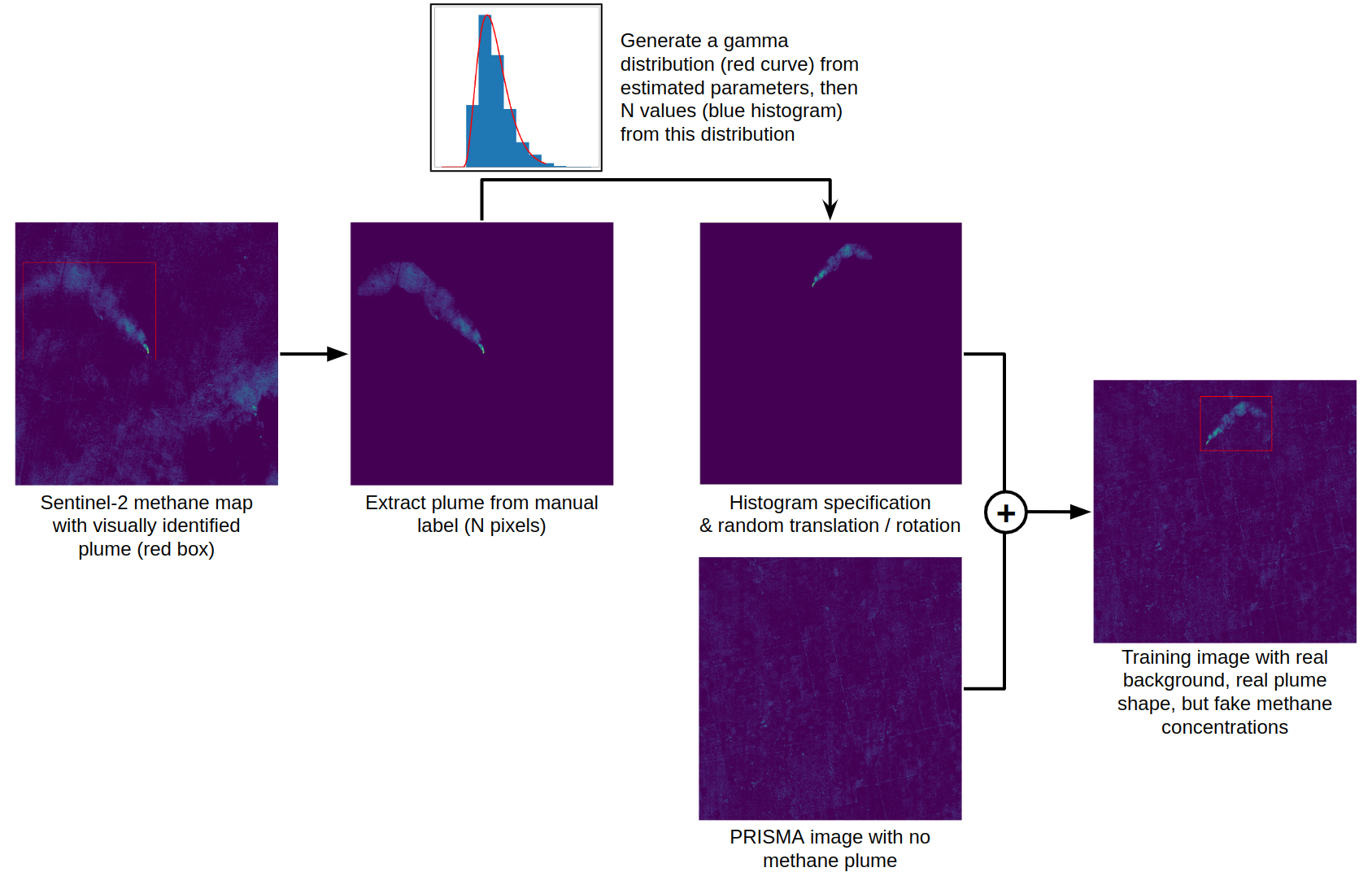}}
    \caption{\centering Protocol for realistic methane plume transfer from a Sentinel-2 image to a PRISMA image with no plume}
    \label{plume_transfer}
\end{figure}

XCH$_4$ plumes exhibit similar spatial distributions driven by atmospheric advection and diffusion: a localized enhancement near the location of the source, and an extended area aligned with the direction of the local wind (advection) decreasing with distance (diffusion). The XCH$_4$ values inside a given plume follow a specific distribution directly dependent on the local atmospheric conditions (wind speed, turbulence conditions, and surface properties). Based on this observation, we assumed a gamma distribution model for the distribution of XCH$_4$ enhancements from the labeled plumes in PRISMA images. At data generation time, for a given Sentinel-2 plume from Ehret~\textit{et al.}~\cite{ehret} with $n_{pix}$ pixels, we sample a random distribution of XCH$_4$ concentrations from the estimated distributions of the parameters. From the gamma distribution defined by these parameters, we can then sample $n_{pix}$ values and apply histogram specification \cite{image_processing} to replace the old values of the plume by these new ones. The final step consists in adding the forged plume to a PRISMA image with no detected XCH$_4$ plume. Addition is preferred to replacement of the concerned pixels in order to preserve the level of noise and the contribution of the underlying elements to the result of the matched-filter. This procedure, illustrated in Fig.~\ref{plume_transfer}, allows us to generate a great amount of unseen training data for the model: any plume from Sentinel-2 among the 1000 labeled samples can be transferred to any of the 150 available backgrounds from PRISMA at a random position and orientation. We also added a criterion on the signal-to-background ratio in order to control the level of contrast of the synthetic plume to the background: at the beginning of the training, we impose that images have a high contrast with the background in order to facilitate model learning. We gradually increase the complexity by lowering this criterion.

\subsubsection{Model}
The detection step is performed using a U-Net architecture, which is one of the most popular architecture for segmentation \cite{unet}. The network is designed to predict a probability map by applying a normalized exponential function (softmax) activation to the output of the final convolutional layer. Each pixel is thus assigned a probability of belonging to a XCH$_4$ plume. This map is then converted into a binary mask by applying a hysteresis thresholding with a low threshold and a large threshold. This approach allows us to detect the whole plume and not only the area close to the source exhibiting the highest XCH$_4$ concentrations. In order to validate the artificial plume generation technique, we train the Convolutional Neural Network (CNN) solely on synthetic data, following the procedure described in Sec.~\ref{data}, and evaluate its performances on XCH$_4$ images from PRISMA. We also compare the performances of our approach to a neural network using an identical architecture trained on the original Sentinel-2 images, with a final transfer learning step to adapt the task to detection in PRISMA images. More precisely, we froze the pre-trained weights of the encoding layers responsible for the extraction of general features and the creation of a meaningful lower dimension representation, and only trained the decoding layers that use this representation for the objective task.

\section{Results}

\paragraph{Methane Retrieval}The implemented retrieval method provides maps of XCH$_4$ enhancements in which plumes can be identified. Examples of such images are available in Fig.~\ref{plume_examples} to illustrate the diversity of data discussed previously that makes detection challenging. We also observe the presence of false positives in the retrieval results. False positives are elements in the scene with a high response but do not correspond to higher XCH$_4$ concentrations. False positives are caused by aerosols or heterogeneous surface properties with strong spectral signatures~\cite{mag1c} similar to methane, such as hydrocarbon paints on buildings or roads, mountain ridges/slopes or sand dunes.

\paragraph{Automatic Plume Detection}

\begin{figure}[h!]
\begin{center}
% \begin{minipage}{1.\textwidth}
% \centering \includegraphics[width = 12.5cm]{images/detection_1.png}
% \end{minipage}
% \hfill
% \begin{minipage}{1.\textwidth}
% \centering \includegraphics[width = 13.5cm]{images/detection_2.png}
% \end{minipage}
% \hfill
% \begin{minipage}{1.\textwidth}
% \centering \includegraphics[width = 12.cm]{images/detection_3.png}
% \end{minipage}
\begin{minipage}{1.\textwidth}
\centering \includegraphics[width = 12.cm]{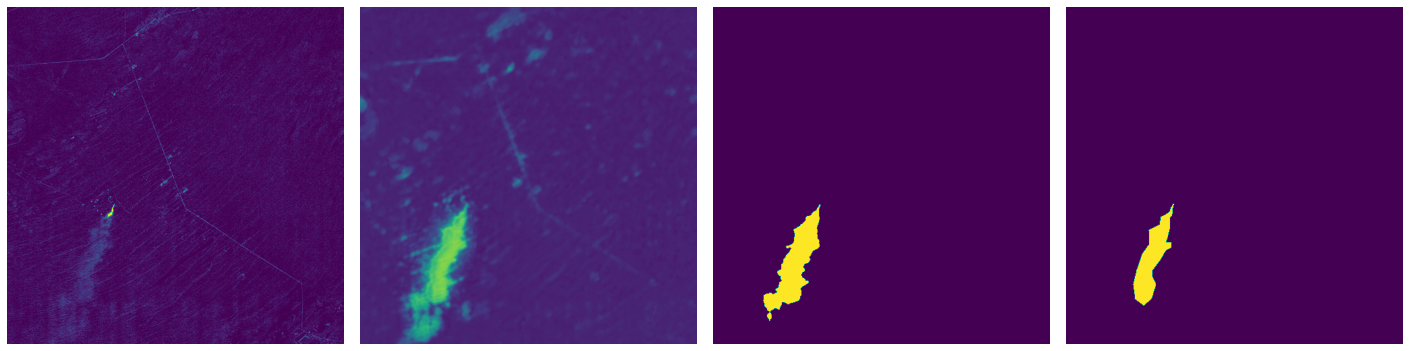}
\end{minipage}
\end{center}
\caption{\centering Examples of automatically-detected XCH$_4$ plumes including XCH$_4$ concentration maps (left column), the probability map predicted by the network (second column), hysteresis thresholding (third column), and manually labeled ground truth (right column)}
\label{detection_res}
\end{figure}

\begin{figure}[h]
\begin{center}
\begin{minipage}{0.45\linewidth}
\centering \includegraphics[width = 3.5cm]{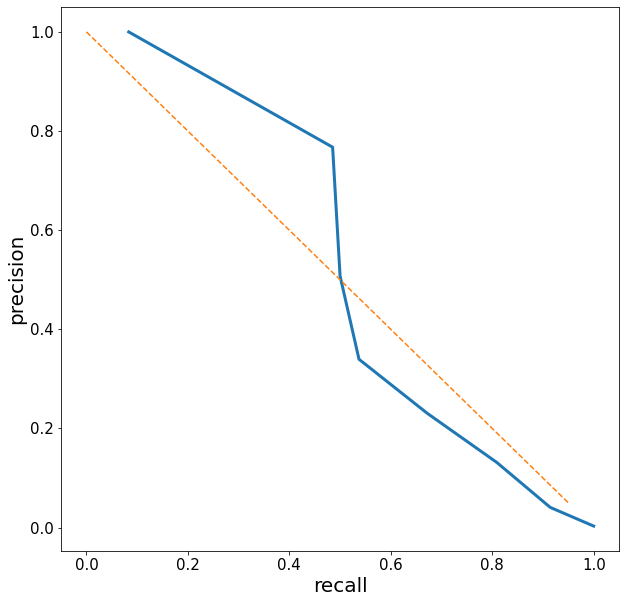}
\caption{\centering detection precision against recall}
\label{tradeoff_recall}
\end{minipage}
\hfill
\begin{minipage}{0.45\linewidth}
\centering \includegraphics[width = 3.5cm]{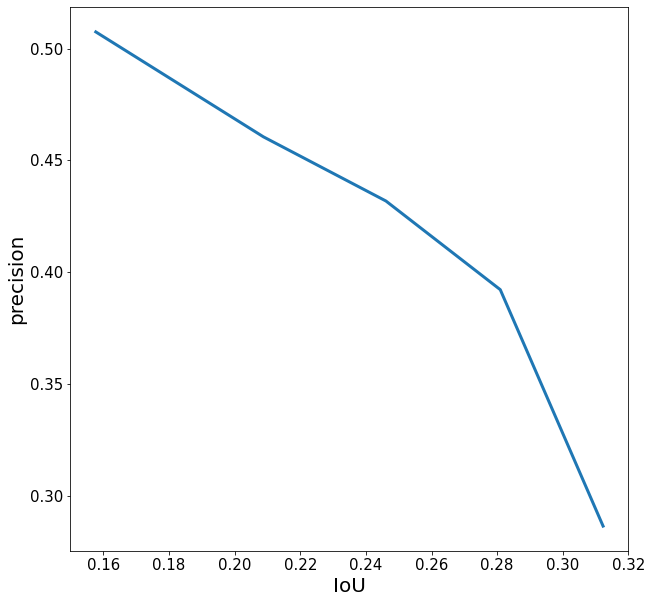}
\caption{\centering detection precision against Intersection over Union (IoU)}
\label{tradeoff_iou}
\end{minipage}
\end{center}
\end{figure}

We present examples of detection results of the network trained solely on synthetic plumes in Fig.~\ref{detection_res}. From a qualitative point of view, the network predicts high probabilities on the whole plume, with a confidence peak on the emission source. It also shows robustness against size variations, even if a few false positives remain, mostly inherited from false positives of the retrieval method. Adjusting the thresholds of the hysteresis postprocessing allows to control a trade-off on the detection performance (see Fig.~\ref{tradeoff_recall}). Indeed, increasing the large threshold reduces the number of false positives, but true plumes may be ignored. Similarly, increasing the low threshold improves the overall Intersection over Union (IoU), but leads to the loss of diffuse parts of the plume (see Fig.~\ref{tradeoff_iou}).

In table~\ref{models_perf}, we make a quantitative comparison between the performance of the model trained from scratch on artificial data, and the model obtained by transfer learning from pre-trained weights learnt on Sentinel-2 images. For both approaches, we consider the hysteresis thresholds producing the best IoU. The detection metrics are computed on a mask basis, a mask being considered a true positive if it intersects a ground-truth mask. The model trained from scratch on synthetic plumes outperforms the model by transfer learning on both the detection and segmentation tasks. The latter notably detects a large number of false positives, leading to a poor precision and IoU even if it reaches a slightly better recall. We also observe a drop when passing from the IoU to the mean IoU (mIoU), which can be explained by the fact that the model often fails to detect the smallest plumes.

% \begin{table}[h!]
%   \caption{Models performance comparison for automatic methane plume detection}
%   \label{models_perf}
%   \centering
%   \begin{tabular}{l|c|c|c|c|c}
%     & \multicolumn{3}{c|}{detection metrics} & \multicolumn{2}{c}{segmentation metrics}\\
%     \hline\hline
%     & precision & recall & f1-score & IoU & mIoU\\
%     \hline
%     Transfer Learning & 0.72 & 0.39 & 0.51 & 0.17 & 0.08\\
%     \hline
%     Plumes Transfer & 0.88 & 0.42 & 0.57 & 0.61 & 0.19
%   \end{tabular}
% \end{table}

\begin{table}[h!]
  \caption{Models performance comparison for automatic methane plume detection}
  \label{models_perf}
  \centering
  \begin{tabular}{l|c|c|c|c|c}
    & \multicolumn{3}{c|}{detection metrics} & \multicolumn{2}{c}{segmentation metrics}\\
    \hline\hline
    & precision & recall & f1-score & IoU & mIoU\\
    \hline
    Transfer Learning & 0.28 & 0.53 & 0.37 & 0.21 & 0.13\\
    \hline
    Plumes Transfer & 0.88 & 0.42 & 0.57 & 0.61 & 0.19
  \end{tabular}
\end{table}

\section*{Conclusion}
In this study, we presented a full processing pipeline for the identification of methane plumes in hyperspectral images from PRISMA. It makes use of classic methods for the spectral recalibration and methane concentrations retrieval. We also propose an automatic detection approach based on a CNN that is trained from scratch using a plume transfer method to generate training samples from methane plumes in Sentinel-2 images. This novel approach allows to train a dedicated model for a new remote sensing technology, while mostly relying on data from previous satellites.

% Further improvements of the model will include an additional RGB or panchromatic image to help the model discard false positives caused by spectral contamination, visible in such images unlike methane.

\newpage
\bibliographystyle{ieeetr}
\bibliography{biblitex}

\newpage
\section*{\centering Appendix}

\hspace{0pt}
\vfill
\vfill
\hspace{0pt}

\newpage

\newpage

\begin{figure}[h!]
\begin{center}
\begin{minipage}{0.31\linewidth}
\centering \includegraphics[width = 4.9cm]{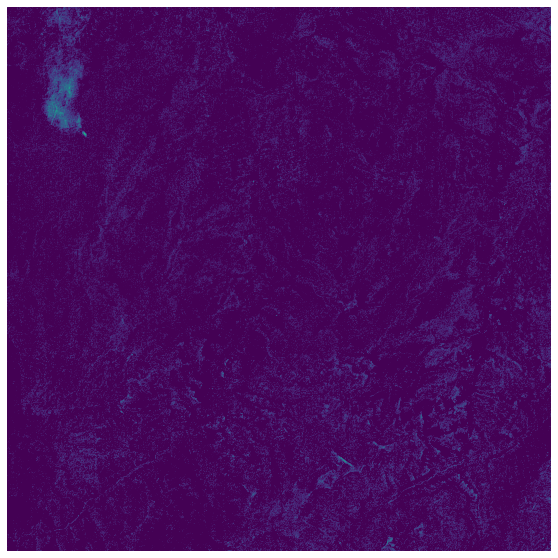}
\end{minipage}
\hfill
\begin{minipage}{0.31\linewidth}
\centering \includegraphics[width = 4.9cm]{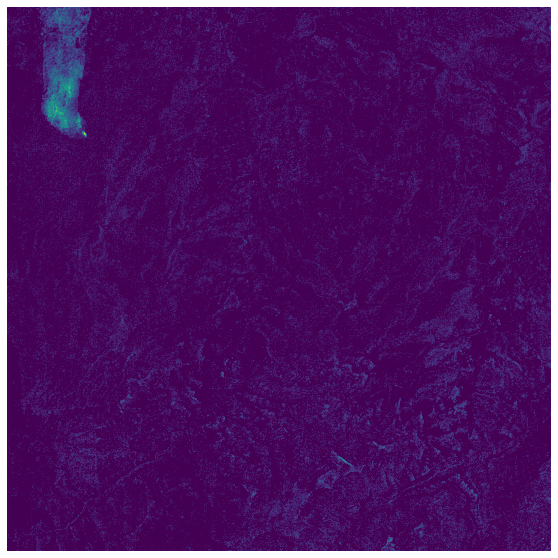}
\end{minipage}
\hfill
\begin{minipage}{0.31\linewidth}
\centering \includegraphics[width = 4.9cm]{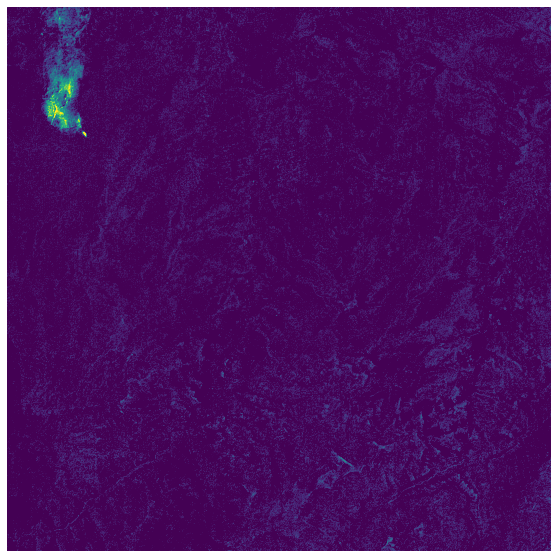}
\end{minipage}
\end{center}
\caption{\centering A synthetic plume added to a PRISMA background with an increasing contrast (from left to right)}
\end{figure}

\begin{figure}[h!]
\begin{center}
\begin{minipage}{0.31\linewidth}
\centering \includegraphics[width = 4.9cm]{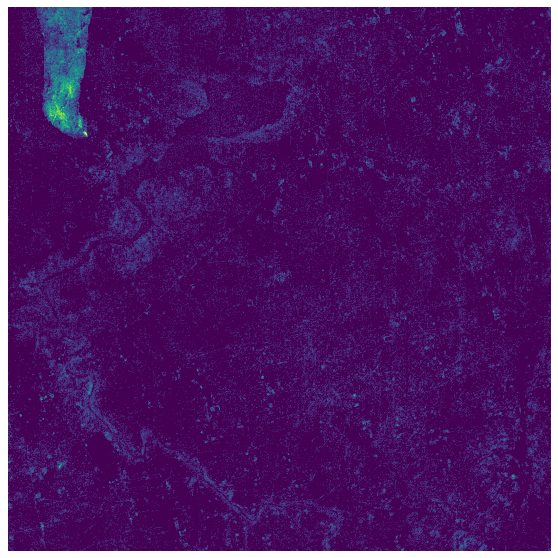}
\end{minipage}
\hfill
\begin{minipage}{0.31\linewidth}
\centering \includegraphics[width = 4.9cm]{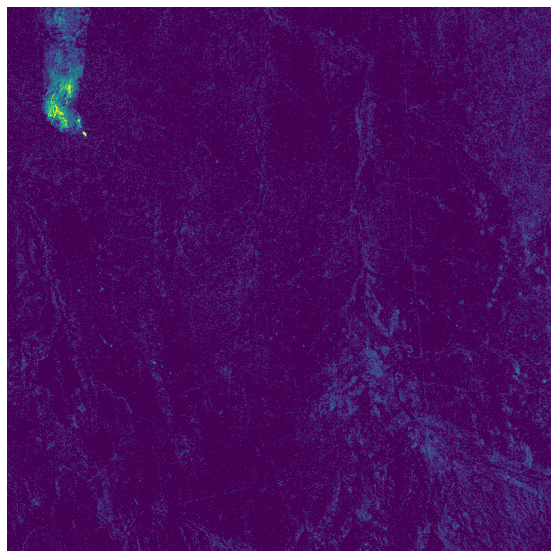}
\end{minipage}
\hfill
\begin{minipage}{0.31\linewidth}
\centering \includegraphics[width = 4.9cm]{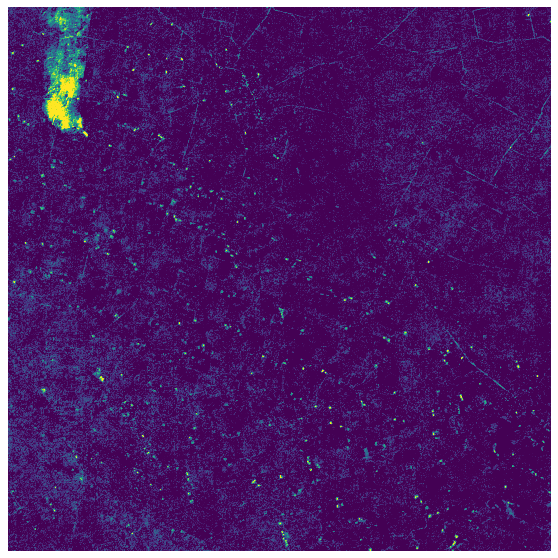}
\end{minipage}
\end{center}
\caption{Same synthetic plume added to different backgrounds}
\end{figure}

\begin{figure}[h!]
\begin{center}
\begin{minipage}{0.31\linewidth}
\centering \includegraphics[width = 4.9cm]{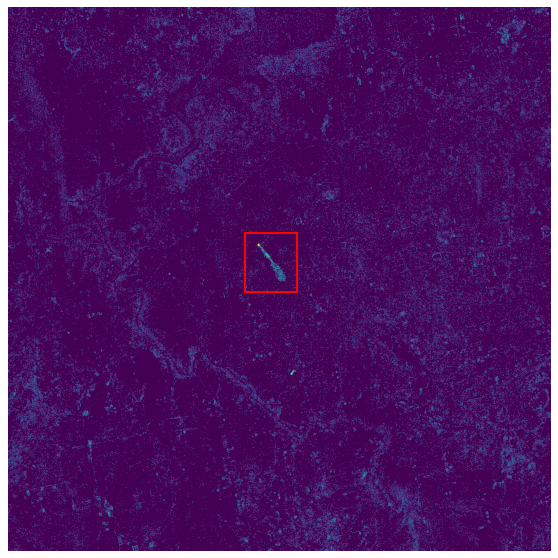}
\end{minipage}
\hfill
\begin{minipage}{0.31\linewidth}
\centering \includegraphics[width = 4.9cm]{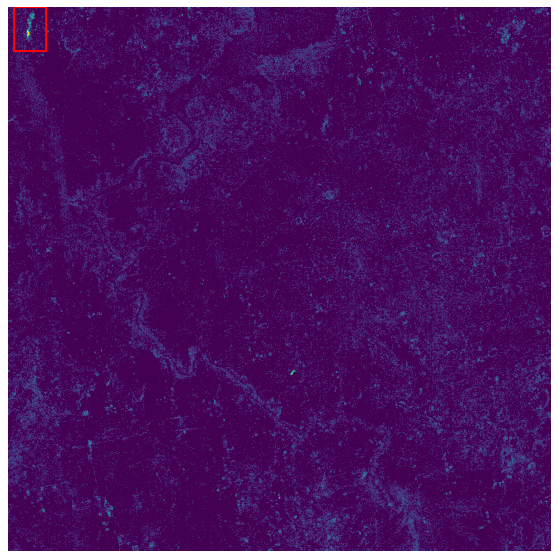}
\end{minipage}
\hfill
\begin{minipage}{0.31\linewidth}
\centering \includegraphics[width = 4.9cm]{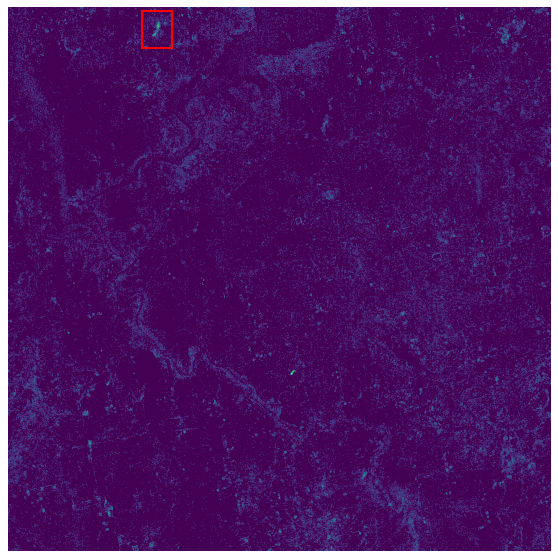}
\end{minipage}
\end{center}
\caption{Several synthetic plumes added to the same background}
\end{figure}

\end{document}